\documentclass[10pt,twocolumn,letterpaper]{article}
\pdfoutput=1
\usepackage[pagenumbers]{cvpr} % To force page numbers, e.g. for an arXiv version

\usepackage{times}
\usepackage{epsfig}
\usepackage{graphicx}
\usepackage{amsmath}
\usepackage{amssymb}
\usepackage{booktabs}
\usepackage{mathrsfs}
\usepackage{bm}

\usepackage{algorithm, algorithmic}
\usepackage{mathdots}
\usepackage{multirow}
\usepackage{microtype} 
\usepackage{array}
\usepackage{setspace}
\usepackage{threeparttable}

\usepackage{graphicx}
\usepackage{diagbox}
\usepackage{wrapfig}
\usepackage{float}
\usepackage{bbm}
\usepackage{url}

\usepackage{bbding}

\usepackage[dvipsnames]{xcolor}

\definecolor{cvprblue}{rgb}{0.21,0.49,0.74}
\usepackage[pagebackref,breaklinks,colorlinks,citecolor=cvprblue]{hyperref}

\def\paperID{9} % *** Enter the Paper ID here
\def\confName{CVPR}
\def\confYear{2024}

\title{Joint Physical-Digital Facial Attack Detection Via Simulating Spoofing Clues}

\author{
	Xianhua He$^{\rm 1}$, 
	Dashuang Liang$^{\rm 1}$, 
	Song Yang$^{\rm 1}$, 
	Zhanlong Hao$^{\rm 1}$\\
	Hui Ma$^{\rm 2}$, 
	Binjie Mao$^{\rm 1}$, 
	Xi Li$^{\rm 1}$, 
        Yao Wang$^{\rm 1}$, 
        Pengfei Yan$^{\rm 1}$,
        Ajian Liu$^{\rm 3}$\thanks{Corresponding author.} \\
        $^{\rm 1}$Vision AI Department, Meituan;
        $^{\rm 2}$M.U.S.T, Macau; %Macau University of Science and Technology, Macau \\
	$^{\rm 3}$MAIS, CASIA, China\\ % Institute of Automation, Chinese Academy of Sciences. 95 
	\tt\footnotesize
	$^1$hexianhua@meituan.com,
 	\tt\footnotesize
	$^3$ajian.liu@ia.ac.cn
}

\begin{document}
\maketitle
\begin{abstract}
Face recognition systems are frequently subjected to a variety of physical and digital attacks of different types. Previous methods have achieved satisfactory performance in scenarios that address physical attacks and digital attacks, respectively. However, few methods are considered to integrate a model that simultaneously addresses both physical and digital attacks, implying the necessity to develop and maintain multiple models. To jointly detect physical and digital attacks within a single model, we propose an innovative approach that can adapt to any network architecture. Our approach mainly contains two types of data augmentation, which we call Simulated Physical Spoofing Clues augmentation (SPSC) and Simulated Digital Spoofing Clues augmentation (SDSC). SPSC and SDSC augment live samples into simulated attack samples by simulating spoofing clues of physical and digital attacks, respectively, which significantly improve the capability of the model to detect ``unseen" attack types. Extensive experiments show that SPSC and SDSC can achieve state-of-the-art generalization in Protocols 2.1 and 2.2 of the UniAttackData dataset, respectively. Our method won first place in ``Unified Physical-Digital Face Attack Detection" of the 5th Face Anti-spoofing Challenge@CVPR2024. Our final submission obtains 3.75\% APCER, 0.93\% BPCER, and 2.34\% ACER, respectively. Our code is available at~\url{https://github.com/Xianhua-He/cvpr2024-face-anti-spoofing-challenge}.
\end{abstract}

% To address the problem of detecting physical attacks across domains, we proposed SPSC to simulate the spoofing clues of physical attacks and improve the ability to detect physical attacks. To address the problem of detecting digital attacks across domains, we proposed SDSC to simulate the spoofing clues of digital attacks and improve the ability to detect digital attacks.     
\section{Introduction}
\label{sec:intro}
Facial recognition systems are widely used in our daily lives, but they are suffering from an increasing number of various types of attacks. The types of facial attacks can be mainly divided into physical attacks and digital attacks. The widely used types of high-frequency physical attacks are: Print attacks \cite{zhang2019dataset,zhang2020casia,liu2019multi}, Replay attacks \cite{liu2021casia,liu2021cross}, and 3D mask attacks \cite{fang2023surveillance,jia2020survey,liu2022contrastive,liu20213d}. Meanwhile, we can briefly summarize digital attacks as digital forgery attacks \cite{rosberg2023facedancer,hong2022depth,wang2021one,wang2021safa,chen2020simswap,heusch2020deep} and adversarial attacks \cite{duan2021advdrop,rony2021augmented,wang2021demiguise,luo2022frequency,yan2022ila,zou2022making}. When addressing physical attacks, central difference convolutions \cite{yu2020searching} and other methods \cite{feng2020learning, yu2021multiple,li2021dynamic,liu2022disentangling,wang2024multi,liu2024cfplfas} can learn low-level and high-level discriminative visual features and achieve advanced performance in classifying live and physical attacks. When detecting digital attacks, deepfake methods \cite{guo2023controllable,shiohara2022detecting,kim2021fretal, zhuang2022uia} can distinguish live and digital forgery attacks by dividing the characteristic subspace of different forgery attack types, and achieve advanced generalization in different digital forgery datasets.
\begin{figure}[t]
   \includegraphics[width=\linewidth]{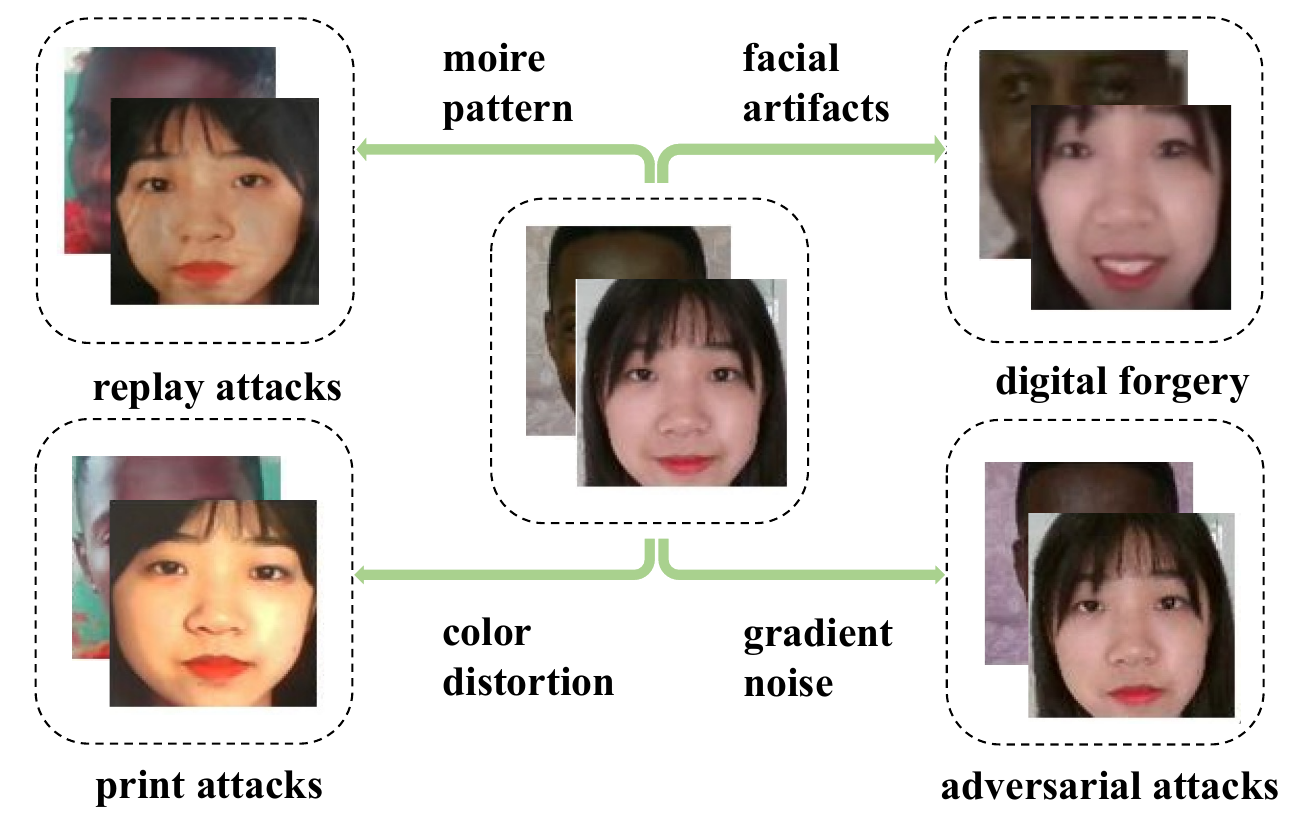}
   \caption{\textbf{Spoofing clues for four attack types.} For replay attacks, print attacks, digital forgeries, and adversarial attacks, the spoofing clues distinct from live samples are identified as moire patterns, color distortion, facial artifacts, and gradient noise, respectively.}
   \vspace{-0.5cm}
   \label{fig:spoofing clues}
\end{figure}

However, few studies detect physical attacks and digital attacks simultaneously. Previous methods separately train models that detect physical and digital attack types, which means that multiple different models need to be deployed to comprehensively judge the final results. This requires more computing resources, and it is more difficult to develop and maintain multiple models. To address this issue, we discuss based on the UniAttackData dataset \cite{fang2024unified} that jointly detects physical attacks and digital attacks. The UniAttackData dataset contains 1800 subjects from 3 different races, covering two types of physical attacks, six types of digital forgery, and six types of adversarial attacks. The same face ID covers all attack types. The UniAttackData dataset defines two protocols. Protocol 1 is designed to evaluate the ability of the model to jointly detect physical attacks and digital attacks. However, there are huge intra-class differences between physical attacks and digital attacks, which brings monumental challenges to the design of the algorithm. Protocol 2 is employed to evaluate the model's ability to detect ``unseen" attack types. The test set for Protocol 2.1 exclusively comprises physical attacks that were not present in the training and development sets. Similarly, the test set for Protocol 2.2 is strictly composed of digital attacks absent from the training and development phases.\\
\indent We discuss the issue of jointly detecting physical and digital attacks and derive some of the following insights. The spoofing clues can improve the model's ability to distinguish between live and various types of attacks. As shown in Figure \ref{fig:spoofing clues}, the spoofing clues typically manifested in physical attacks, such as print and replay attacks, are color distortions and moire patterns. In terms of digital forgery attacks, including face-swapping or face generation, the prevalent spoofing clues are facial artifacts or distortions. Additionally, adversarial attacks frequently introduce spoofing clues by adding specified gradient noise to the original image.\\
\indent Motivated by the discussions above, we introduce SPSC and SDSC to improve the model's ability to detect physical attacks and digital attacks, respectively. Specifically, in Protocol 2.1, we use SPSC to simulate color distortion and moire patterns and augment the live samples into physical attack samples for training. In Protocol 2.2, we use SDSC to simulate the artifacts caused by face swapping and augment the live samples into digital attack samples for training. SPSC and SDSC improve the ability of the models in protocols 2.1 and 2.2 to detect ``unseen" attack types, respectively. Meanwhile, using SPSC and SDSC in Protocol 1 also improves the generalization performance of the model.\\
\indent In summary, the main contributions of this paper are summarized as follows:
% \vspace{-0.1cm}
\begin{itemize}
\setlength{\itemsep}{2pt}
\item We propose Simulated Physical Spoofing Clues augmentation (SPSC) to simulate the spoofing clues of physical attacks and address the issue of cross-attack type detection from digital to physical attacks in protocol 2.1.
\item 
We introduce Simulated Digital Spoofing Clues augmentation (SDSC) to simulate the spoofing clues of digital attacks and overcome the barriers to how physical attacks generalize to digital attacks in protocol 2.2.
\end{itemize}

% address the issue of how physical attacks generalize to digital attacks in protocol 2.2.

% We introduce Simulated Physical Spoofing Clues Augmentation (SPSC) to mimic physical attacks and facilitate digital-to-physical cross-attack type detection in protocol 2.1, utilizing both ColorJitter and moiré pattern augmentation.

% The development of Simulated Digital Spoofing Clues Augmentation (SDSC) to emulate digital attack indicators and tackle the physical-to-digital attack generalization challenge in protocol 2.2.

\begin{figure*}[t]
   \includegraphics[width=\linewidth]{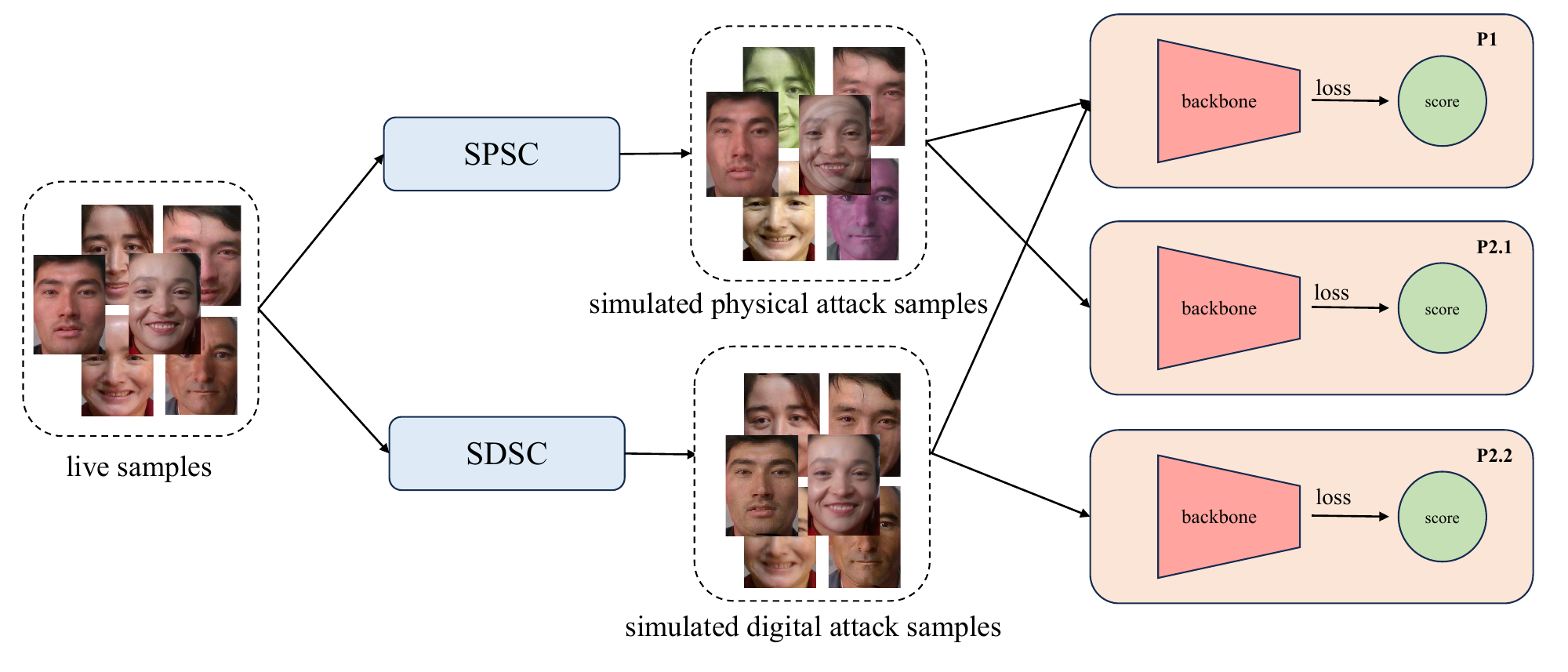}
   \caption{\textbf{The overview pipeline of our method.} We propose Simulated Physical Spoofing Clues augmentation (SPSC), which augments live samples into simulated physical attack samples for training within protocols 1 and 2.1. Concurrently, we present Simulated Digital Spoofing Clues augmentation (SDSC), converting live samples into simulated digital attack samples, tailored for training under protocols 1 and 2.2.}
   % \vspace{-0.5cm}
   \label{fig:pipeline}
\end{figure*}

% We introduce simulate physical spoofing clues augmentation(SPSC) and simulate digital spoofing clues augmentation(SDSC) for physical attacks and digital attacks, respectively. The live samples is converted into simulated attack samples through SPSC or SDSC which is used in different protocols, respectively.

\section{Related work}
% After years of research and development, a large amount of work has been accumulated on the face anti-spoofing (FAS) task against physical attacks. 
\subsection{Physical Attack Detection.}
With the development of deep learning, convolutional neural networks (CNN) have gradually become the mainstream method for solving the task of FAS. Liu et al.~\cite{Liu2018Learning} utilize physical-based depth information as a supervisory signal instead of binary classification loss. Yu et al.~\cite{yu2020searching} propose Central Difference Convolution (CDC), which is able to capture intrinsic detailed patterns via aggregating both intensity and gradient information. Feng et al.~\cite{feng2020learning} propose a residual-learning framework to learn the discriminative live-spoof differences, which are defined as the spoof cues. PIFAS~\cite{liu2022disentangling} decomposes faces into appearance information and pose codes to capture liveness and liveness-unrelated features, respectively. AA-FAS~\cite{Liu_2023_CVPR} regards FAS as a unified framework with the attack and defense systems, which employs adversarial training to optimize the defense system against unseen attacks. Although these algorithms have achieved astonishing results in intra-datasets experiments, their performance deteriorates severely when faced with unknown domains. To solve these limits, Domain Generalization (DG) based methods~\cite{shao2019multi,liu2021adaptive,Chen2021a,srivatsan1} can conquer this by taking advantage of multiple source domains without seeing any target data. Jia et al.~\cite{jia2020single} propose an end-to-end single-side domain generalization framework (SSDG) to improve the generalization ability of face anti-spoofing. Sun at al.~\cite{sun2023rethinking} encourages domain separability while aligning the live-to-spoof transition (i.e., the trajectory from live to spoof) to be the same for all domains. Huang et al.~\cite{huang2022adaptive} introduce the ensemble adapters module and feature-wise transformation layers in the ViT to adapt to different domains for robust performance with a few samples. IADG~\cite{zhou2023instance} framework aligns features on the instance level, reducing sensitivity to instance-specific styles. MDIL~\cite{wang2024multi} consists of an adaptive domain-specific experts (ADE) framework based on the vision transformer, and an asymmetric classifier is designed to keep the output distribution of different classifiers consistent. CFPL-FAS~\cite{liu2024cfplfas} makes use of large-scale VLMs like CLIP and leverages the textual feature to dynamically adjust the classifier's weights for exploring generalizable visual features. 

Multi-modal FAS has gained significant attention due to the increasing sophistication of high-quality attacks. However, these multi-modal fusion-based algorithms require the testing phase to provide the same modal types as the training phase, severely limiting their deployment scenarios. Flexible modality-based methods~\cite{liu2021face,ijcai2022p165,yu2023visual,yu2023flexiblemodal,liu2023fm} aim to improve the performance of any single modality by leveraging available multi-modal data. 
% CMA-FAS~\cite{liu2021face} closes the visible gap between different modalities via a generative model that maps inputs from one modality to another. MA-ViT~\cite{ijcai2022p165} adopts the early fusion to aggregate all the available training modalities data and enables flexible testing of any given modal samples. MA-ViT develops the Modality-Agnostic Transformer Block in MA-ViT to eliminate modality-related information for each modal sequence and supplement modality-agnostic liveness features from another modal sequence, respectively. FM-ViT~\cite{liu2023fm} retains a specific branch for each modality to capture different modal information and introduces the Cross-Modal Transformer Block to guide each modal branch to mine potential features from informative patch tokens and to learn modality-agnostic liveness features by enriching the modal information of its own CLS token, respectively.  

% \cite{afchar2018mesonet, haliassos2021lips, kim2021fretal, masi2020two, nguyen2019multi, zhuang2022uia}.

\subsection{Digital Attack Detection.}
% Face forgery detection based on convolutional neural network (CNN) has been widely used. Some methods attempt to learn common features across different forgery domains. For example, 
Some work in \cite{nadimpalli2022improving} formulate deep fake detection as a hybrid combination of supervised and reinforcement learning (RL). \cite{nadimpalli2022improving} chooses the top-k augmentations for each test sample by an RL agent in an image-specific manner and the classification scores, obtained using CNN, of all the augmentations of each test image are averaged together for final real or fake classification. Guide-Space \cite{guo2023controllable} is a controllable guide-space method to enhance the discrimination of different forgery domains. The well-designed guide space can simultaneously achieve both the proper separation of forgery domains and the large distance between real-forgery domains in an explicit and controllable manner. Self-Blending~\cite{shiohara2022detecting} presents novel synthetic training data to detect deepfakes. Self-Blending is generated by blending pseudo source and target images from single pristine images, reproducing common forgery artifacts (e.g., blending boundaries and statistical inconsistencies between source and target images).

\begin{figure*}[t]
     \centering
     \includegraphics[width=\linewidth]{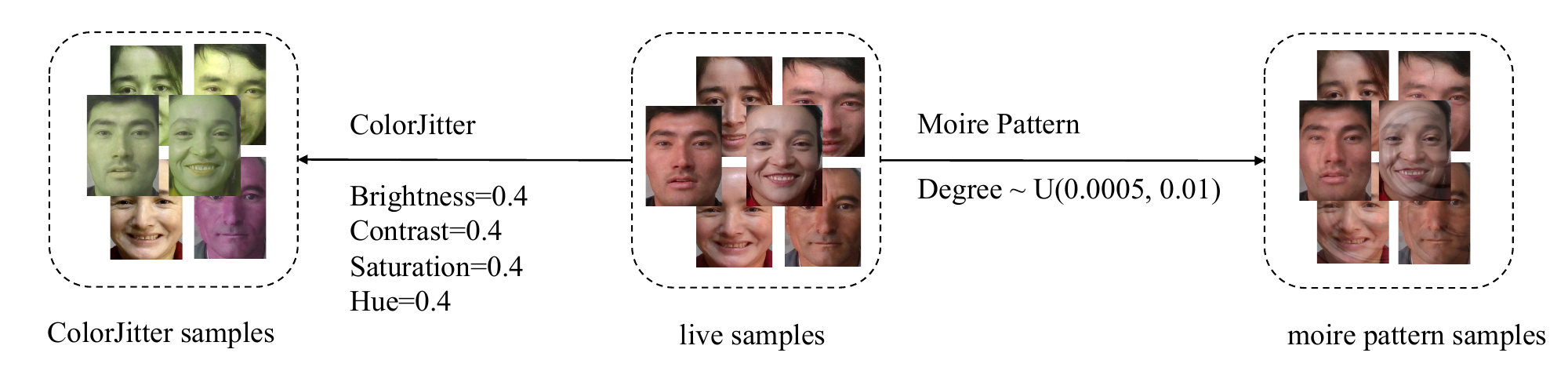}
     \caption{Live samples simulate print attacks through varying degrees of ColorJitter and replay attacks through varying degrees of moire pattern augmentation.}
     \label{fig:spsc}
    % \vspace{-0.5cm}
\end{figure*}

\section{Methodology}
\label{sec:method}
In this section, we introduce an overview of our method in Section \ref{subsec:3.1}. Subsequently, we elaborate on Simulated Physical Spoofing Clues augmentation and Simulated Digital Spoofing Clues augmentation in Section \ref{subsec:3.2} and Section \ref{subsec:3.3}, respectively.

\subsection{Overview}
\label{subsec:3.1}
As depicted in Figure \ref{fig:pipeline}, our method principally introduces two targeted data augmentation strategies, designated as Simulated Physical Spoofing Clues augmentation (SPSC) and Simulated Digital Spoofing Clues augmentation (SDSC). We augment live samples into simulated attack instances via SPSC or SDSC, subsequently extracting features through a neural network and computing the Cross-Entropy loss to refine classification networks. The augmentations of SPSC and SDSC can be seamlessly incorporated into additional frameworks. The inference phase is consistent with established baseline methods, allowing straightforward score determination. Under Protocol 1, both SPSC and SDSC data augmentations are employed. For protocol 2.1, the model needs to be generalized to detect ``unseen" physical attacks, and only SPSC is used to enhance the model's detection performance against physical attacks. For protocol 2.2, the objective is for the model to generalize to ``unseen" digital attack modalities; hence, only SDSC data augmentation is implemented to improve the model's digital attack detection capabilities.
% As shown in Figure \ref{fig:pipeline}, our method mainly proposes two targeted data augmentation approaches, namely Simulated Physical Spoofing Clues augmentation (SPSC) and Simulated Digital Spoofing Clues augmentation (SDSC). We transform the live samples into attack samples through SPSC or SDSC, extracting features through the neural network, and calculate Cross-Entropy loss to optimize two classification networks. SPSC and SDSC augmentation can be easily extended to other methods. The inference stage aligns with other baseline methods, in which the score is predicted directly. For protocol 1, both data augmentation methods, SPSC and SDSC, are used. For protocol 2.1, we need to generalize the model to detect "unseen" physical attack types, so we only need to use SPSC data augmentation to optimize the model's ability to detect physical attacks. For protocol 2.2, we need the model to generalize to detect "unseen" digital attack types, so we only need to use SDSC data augmentation to optimize the model's ability to detect digital attacks.

\subsection{Simulated Physical Spoofing Clues}
\label{subsec:3.2}
Our comprehensive analysis of physical attack characteristics has led to the development of the Simulated Physical Spoofing Clues augmentation (SPSC), which integrates both ColorJitter and moire pattern augmentation. As demonstrated in Figure \ref{fig:spsc}, we simulate print attacks on live samples through a spectrum of ColorJitter adjustments, and similarly, we emulate replay attacks by applying varying degrees of moire pattern augmentation. This approach allows us to simulate the distinct visual artifacts characteristic of each attack type, thereby enriching the robustness of spoofing detection models. \textbf{ColorJitter:} Print presentation attacks frequently leave behind spoofing cues manifesting as color distortions. To emulate these distortions, ColorJitter serves as an effective tool for creating artificial spoofing cues. By applying ColorJitter, a live sample is converted into a simulated attack sample, effectively capturing the essence of a print attack scenario. For this purpose, we calibrate the ColorJitter settings for brightness, contrast, saturation, and hue uniformly to a factor of 0.4. \textbf{Moire Pattern augmentation:} Spoofing cues in replay presentation attacks are often characterized by the presence of moire patterns. To address this issue, we design a moire pattern generation algorithm that effectively simulates these distinctive patterns. The specifics of this algorithm are presented in the form of pseudo-code, as detailed in Algorithm \ref{algorithm:moire}. This algorithmic approach enables us to create nuanced moire effects that closely mirror those found in actual replay attacks. First, you need to obtain the height and width of the original image, calculate the center point of the image, and randomly generate the moire intensity degree, which follows a uniform distribution from 0.0005 to 0.01. Then, create grid coordinates and generate grid coordinates $(X, Y)$ for the $(x, y)$ coordinates of each pixel. Afterward, calculate the offset and polar coordinate parameters, and calculate the $(X, Y)$ offset $(X_{offset}, Y_{offset})$ of each point relative to the center point. Subsequently, calculate the angle $\theta$ and radius $\rho$ in polar coordinates and calculate new $(X, Y)$ coordinates $(X_{new}, Y_{new})$ based on angle and radius adjustments. Finally, limit the coordinate range and map pixels and limit new coordinates $(X_{new}, Y_{new})$ to the image range. Combining the source image with the mapped pixel values generates an image with a moire pattern effect. 

% \textbf{Moire Pattern augmentation:} The spoofing clues based on replay presentation attacks are usually caused by the moire pattern. We designed a moire pattern generation algorithm to simulate the moire pattern generated. The pseudo-code of the algorithm is shown in Algorithm \ref{algorithm:moire}. First, you need to obtain the height and width of the original image, calculate the center point of the image, and randomly generate the moire intensity degree, which follows a uniform distribution from 0.0005 to 0.01. Then, create grid coordinates and generate grid coordinates $(X, Y)$ for the $(x, y)$ coordinates of each pixel. Afterward, calculate the offset and polar coordinate parameters, and calculate the $(X, Y)$ offset $(X_{offset}, Y_{offset})$ of each point relative to the center point. Subsequently, calculate the angle $\theta$ and radius $\rho$ in polar coordinates and calculate new $(X, Y)$ coordinates $(X_{new}, Y_{new})$ based on angle and radius adjustments. Finally, limit the coordinate range and map pixels and limit new coordinates $(X_{new}, Y_{new})$ to the image range. Combining the source image with the mapped pixel values generates an image with a moire pattern effect. 

\begin{algorithm}
\caption{Add Moire Pattern to Image}
\begin{algorithmic}[1]
\REQUIRE $src$: source image
\STATE $height, width \gets \text{dimensions of } src$
\STATE $center \gets (height / 2, width / 2)$
\STATE $degree \gets \text{random value between 0.0005 and 0.01}$
\STATE $x \gets \text{array from 0 to } width - 1$
\STATE $y \gets \text{array from 0 to } height - 1$
\STATE $X, Y \gets \text{meshgrid of } x \text{ and } y$
\STATE $offset_X \gets X - center[0]$
\STATE $offset_Y \gets Y - center[1]$
\STATE $\theta \gets \text{arctan2}(offset_Y, offset_X)$
\STATE $\rho \gets \sqrt{offset_X^2 + offset_Y^2}$
\STATE $new_X \gets center[0] + \rho \cdot \cos(\theta + degree \cdot \rho)$
\STATE $new_Y \gets center[1] + \rho \cdot \sin(\theta + degree \cdot \rho)$
\STATE $new_X \gets \text{clip}(new_X, 0, width - 1)$
\STATE $new_Y \gets \text{clip}(new_Y, 0, height - 1)$
\STATE $dst \gets 0.8 \cdot src + 0.2 \cdot src[new_Y, new_X]$
\RETURN $dst$ as unsigned 8-bit integer
\end{algorithmic}
\label{algorithm:moire}
\end{algorithm}

\begin{figure*}[t]
     \centering
     \includegraphics[width=\linewidth]{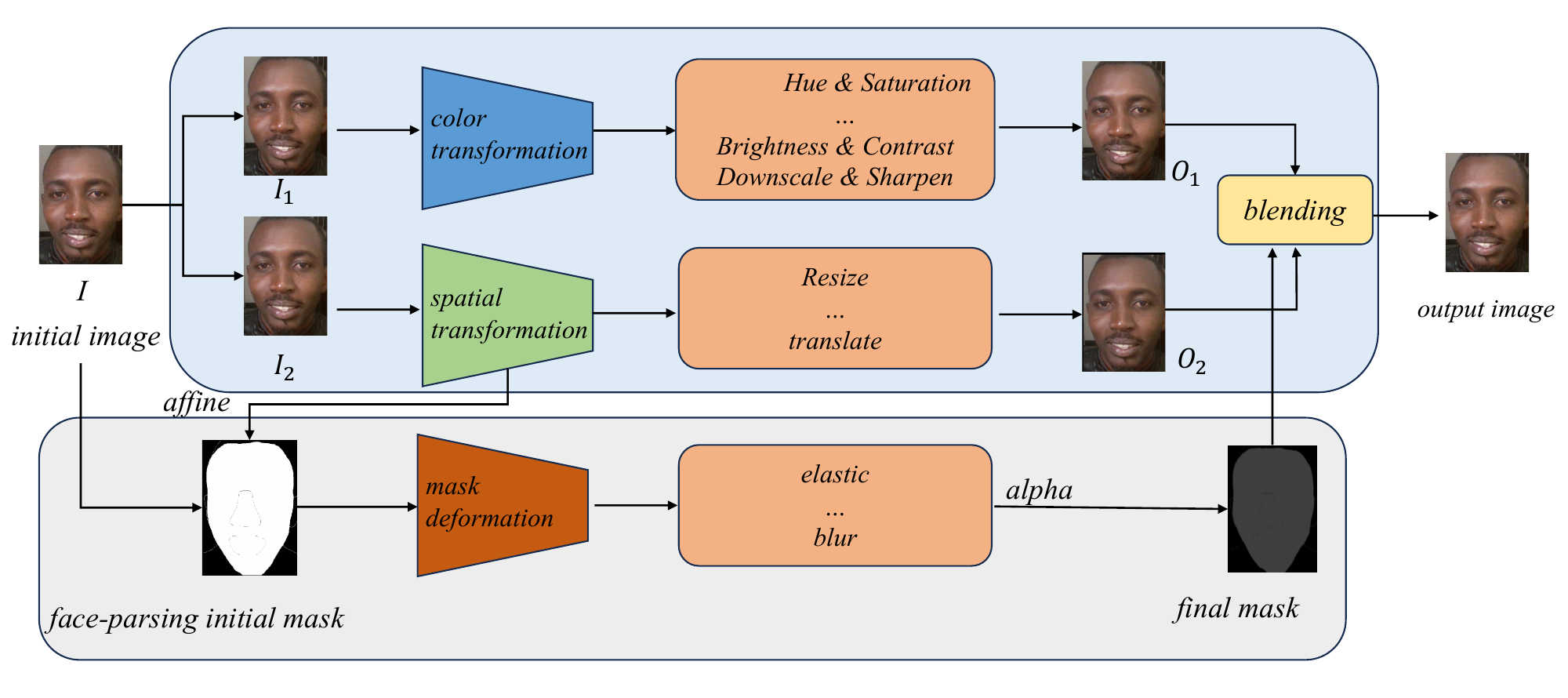}
     \caption{A live sample is transformed into a digital forgery attack sample by Simulated Digital Spoofing Clues augmentation.}
     \label{fig:blending}
    % \vspace{-0.5cm}
\end{figure*}

\subsection{Simulated Digital Spoofing Clues}
\label{subsec:3.3}
Digital forgery attacks often leverage face swapping or face generation algorithms, which typically induce distortions or artifacts within the facial region. Inspired by Self-Blending \cite{shiohara2022detecting}, we propose the Simulated Digital Spoofing Clues augmentation (SDSC). This method is designed to simulate the distortions and artifacts characteristic of digital forgery facial images. As shown in Figure \ref{fig:blending}, the process of SDSC is divided into three steps. \textbf{1)} Obtain pseudo source image and target image for blending. Copy the original Image $I$ to get $I_{1}$ and $I_{2}$. $I_{1}$ is used as a pseudo source image and is augmented by color transformation (e.g., Hue, Brightness, and Downscale) to obtain $O_{1}$. $I_{2}$ is used as a target image and is augmented by spatial transformation (e.g., Resize, translate) to obtain $O_{2}$. The boundary and landmark between $O_{1}$ and $O_{2}$ are misaligned. \textbf{2)} The original image $I$ is segmented through the face parsing algorithm to obtain the face mask. Subsequently, the face mask undergoes affine transformation through spatial transformation, and the final mask is obtained through the augmentation of mask deformation (e.g., elastic, blur). \textbf{3)} $O_{1}$, $O_{2}$ and final mask are blended according to formula \ref{eq: blending} and the forgotten image is output.
\begin{equation}
\label{eq: blending}
   O_{forgery} = O_{1}\odot mask + O_{2}\odot (1-mask)
\end{equation}

% Digital forgery attack methods are usually based on face-swapping algorithms or face-generation algorithms, which usually cause distortion or artifacts in the face area. Inspired by work \cite{shiohara2022detecting}, we introduce Simulated Digital Spoofing Clues augmentation(SDSC) to simulate the distortion and artifacts of the face area. As shown in Figure \ref{fig:blending}, SDSC is divided into three steps. 

\section{Experiments}
\label{sec:experiments}
\subsection{Experimental Settings}
\begin{figure*}[t]
   \includegraphics[width=\linewidth]{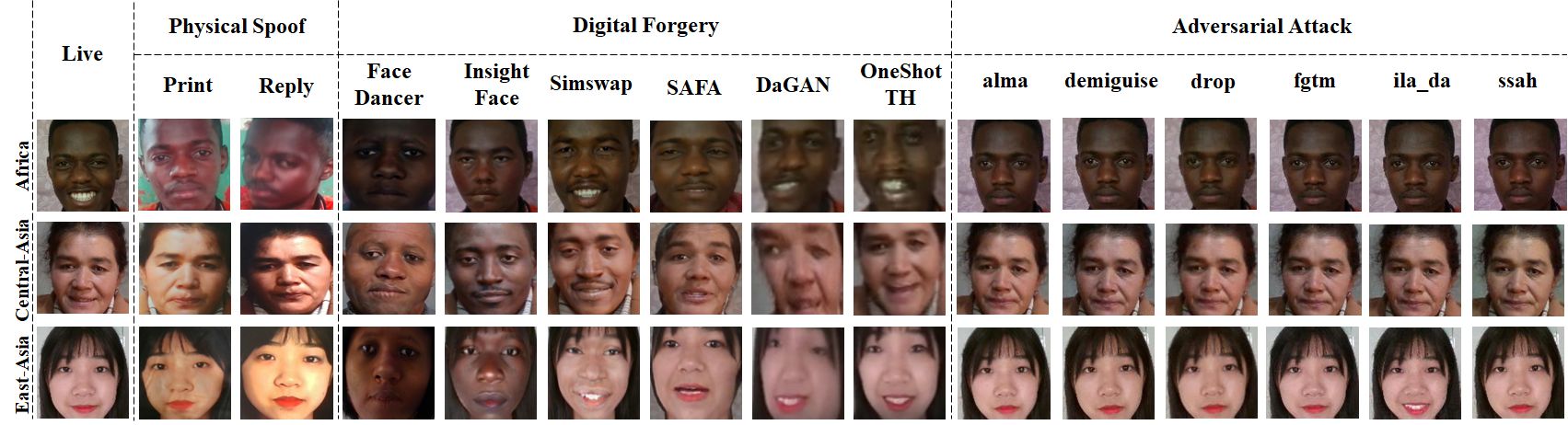}
   \caption{\textbf{The overview of UniAttackData Dataset~\cite{fang2024unified}.} The same face ID forges physical attack videos through two types of physical attacks (print and replay). For every live video, forge digital attacks through six digital editing algorithms and six adversarial algorithms. The attack type of each sample is indicated at the top of the graph.}
   \vspace{-0.5cm}
   \label{fig:dataset}
\end{figure*}

\paragraph{UniAttackData Datase.} 
As shown in Figure \ref{fig:dataset}, the UniAttackData dataset \cite{fang2024unified} expands on the CASIA-SURF \cite{liu2021casia} dataset, featuring 1800 subjects of three races: African, East Asian, and Central Asian. It includes two kinds of physical attack methods (Print and Replay), along with six types of digital forgery attacks and six types of adversarial attacks. UniAttackData defines two protocols to make sure that they can effectively test the system's ability to detect different types of attacks jointly. Protocol 1 is designed to test the system's ability to detect a unified attack type that encompasses both physical and digital attacks. Huge intra-class distance and diverse attacks bring more challenges to algorithm design. Protocol 2 evaluates the model's detection capabilities for ``unseen" attack types, testing algorithmic adaptability across the diverse and unpredictable spectrum of physical and digital attacks. The test set of protocol 2.1 comprises exclusively novel physical attacks, while the test set of protocol 2.2 contains solely unseen digital attacks, both distinct from the training and development datasets.
%Protocol 2 is used to evaluate the model’s ability to detect “unseen” attack types. The large differences and unpredictability between physical and digital attacks challenge the portability of algorithms. The test set of protocol 2.1 contains only physical attacks that have not been seen in the training and development sets. The test set of protocol 2.2 only contains digital attacks that have not been seen in the training and development sets.

\paragraph{Evaluation Metrics.} 
The evaluation protocol used to evaluate performance follows established standards within the field of Face Anti-Spoofing (FAS). Specifically, we utilize the widely accepted metric comprising the Attack Presentation Classification Error Rate (APCER), the Bona Fide Presentation Classification Error Rate (BPCER), and the Average Classification Error Rate (ACER). They can be formulated as:
\begin{equation}
\begin{split}
APCER &= \frac{FP}{FP + TN}, \\
BPCER &= \frac{FN}{FN + TP}, \\
ACER &= \frac{APCER + BPCER}{2},
\end{split}
\end{equation}
where $FP$, $FN$, $TN$, and $TP$ denote the counts of false positive, false negative, true negative, and actual positive instances, respectively. ACER is used to determine the final ranking in the 5th Face Anti-spoofing Challenge@CVPR2024.

\paragraph{Data Preprocess.}
\textbf{Face detection:} For images larger than 700 pixels in width and height, we apply face detection and expand the bounding box by 20 pixels for face cropping. Images without detected faces undergo a center crop to yield a 500$\times$500 input for the model. \textbf{Obtain the face mask:} Face parsing is performed on live samples to generate face masks for SDSC.

\paragraph{Architecture Details.}
Our method can be easily transferred to any backbone network. Since the amount of dataset from the training sample is relatively small, we chose Resnet50 \cite{he2016deep} as the backbone of the classification network. We believe that for this task, the final results obtained by our method will not be significantly different in different backbone networks.

\paragraph{Training Details.} 
We distinguish 3 protocols and train 3 models, respectively. In the training stage, We utilize AdamW optimizer \cite{adamw} to train our model with a learning rate of $1e^{-3}$, and the weight decay is $5e^{-4}$. The cosine learning rate schedule is employed to adjust the learning rate. We resize the image to 224$\times$224 as the input image for model training. The data augmentation of RandomResizedCrop and HorizontalFlip is used during training. We use Cross Entropy loss as the loss function and set different weights to balance the live and attack sample loss. In protocol 1, we used the complete train and dev sets. In protocols 2.1 and 2.2, we use the whole train set and live samples from the development set. We train the model on a single A100(80G) GPU for 200 epochs, with the batch size set to 512. Each protocol requires only one hour to train and one minute to test.

\subsection{Comparison with SOTA Methods}
We compare the performance of our proposed method with state-of-the-art (SOTA) methods (teams) on the UniAttackData dataset \cite{fang2024unified}. Table \ref{table:final_rank} summarizes the results of the comparison of four metrics: AUC, APCER, BPCER, and ACER. In the SOTA method, VAI-Face achieves the lowest BPCER with a value of 0.25\%. However, our proposed method achieves the highest performance on the AUC, APCER and ACER metrics, with values of 99.69\%, 3.75\%, and 2.33\%, respectively. In addition, our method achieves the lowest APCER with a value of 3.75\%, which is significantly lower than the other methods. The experimental results validate the efficacy of our method, with SPSC and SDSC demonstrating impressive performance in detecting ``unseen" attack types under Protocols 2.1 and 2.2.

\begin{table}[t]
\footnotesize
\centering
% \vspace{2mm}
\begin{tabular}{c c c c c}
\toprule
    Team & AUC $\uparrow$ & APCER $\downarrow$ & BPCER $\downarrow$ & ACER $\downarrow$ \\ 
    \midrule
    VAI-Face &  87.75 &  34.00  &  \textbf{0.25}  &  17.13 \\
    BSP-Idiap &  96.33  &  23.06  &  9.36  &  16.22 \\
    duileduile &  98.68  &  5.50 &  5.51 &  5.51 \\
    SeaRecluse &  96.56  & 6.47  &  0.39  &  3.43 \\
    Ours &  \textbf{99.69}  &  \textbf{3.75}  &  0.92  &  \textbf{2.33} \\
    % w/o DIFF &  x &  x &  x \\      
\bottomrule
\end{tabular}
\caption{Comparing results on the test set of the UniAttackData Dataset~\cite{fang2024unified}. Our method achieves the highest performance on the AUC(\%), APCER(\%) and ACER (\%) metrics.}
\label{table:final_rank}
\end{table}
\subsection{Protocol Result} 
We define the baseline method without SPSC and SDSC. Then, we employ our method and retrain three models under the corresponding protocol to compare with the baseline model. The results, as shown in Table \ref{table:protocol}, indicate a marginal improvement of our method over the baseline. For protocol 1, our method has a slight improvement compared to the baseline. For Protocol 2.1, our method achieved a substantial improvement compared to the baseline, reducing the ACER from 38.05\% to 1.32\%. Similarly, in Protocol 2.2, our approach significantly outperformed the baseline, decreasing the ACER from 44.35\% to 1.65\%. By averaging the results across the three protocols, our proposed method reduced the ACER from 27.54\% to 1.06\%, a notable decline of 26.48\%. This highlights the effectiveness of our approach. Finally, the results of our replication were slightly better than those we submitted to the competition organizers, further underscoring the robustness of our method.
\begin{table}[t]
\footnotesize
\centering
% \vspace{2mm}
% \resizebox{0.47\textwidth}{10mm}{
% \begin{tabularx}{0.47\textwidth}{X X X X X X}
\begin{tabular}{c c c c c}
\toprule
    Protocol & Methods & APCER $\downarrow$ & BPCER $\downarrow$ & ACER $\downarrow$ \\ 
    \midrule
    \multirow{2}{*}{P1} 
    &  w/o SPSC\&SDSC & 0.17  &  0.28  &  0.23 \\
    &  w/ SPSC\&SDSC & 0.31  &  0.09  &  0.20 \\
    \midrule
    \multirow{2}{*}{P2.1} 
    &  w/o SPSC\&SDSC & 76.03  &  0.06  &  38.05 \\
    &  w/ SPSC & 2.55  &  0.09  &  1.32 \\
    \midrule
    \multirow{2}{*}{P2.2} 
    &  w/o SPSC\&SDSC & 88.59  & 0.11  & 44.35 \\
    &  w/ SDSC & 1.73  &  1.58  &  1.65 \\
    \midrule
    \multirow{2}{*}{All} 
    &  w/o SPSC\&SDSC & 54.93  &  0.15 & 27.54 \\
    &  w/ SPSC\&SDSC & 1.53 & 0.69 & 1.06  \\

\bottomrule
\end{tabular}
\caption{Compare the results of baseline(w/o SPSC\&SDSC) and ours method(w/ SPSC\&SDSC) on all protocols}
\label{table:protocol}
\end{table}

\subsection{Ablation Study}
\paragraph{Simulated Physical Spoofing Clues} 
SPSC contains ColorJitter and moire pattern augmentations, targeting print and replay attack spoofing cues, respectively. Using the complete training set and live development samples, SPSC significantly outperforms the baseline. As shown in Table \ref{table:spsc}, ColorJitter and moire pattern augmentations individually lower the baseline ACER of 38.05\% to 3.62\% and 6.18\%, respectively. The combined augmentation of SPSC further reduces ACER to 1.32\% in protocol 2.1, a substantial improvement of 36.73\% over the baseline.

% SPSC includes two augmentations: ColorJitter and moire pattern augmentation, which are used to simulate the spoofing clues of print attack and replay attack, respectively. In protocol 2.1, the full train set and live samples in the dev set are used for training, and Resnet50\cite{he2016deep} is used as the backbone to train the baseline result. As presented in Table \ref{table:spsc}, compared to the baseline ACER result of 38.05\%, using ColorJitter and moire pattern augmentation achieved ACER 3.62\% and 6.18\% results, respectively. ACER has been significantly reduced, indicating that these two types of data augmentation are very effective. Finally, we merged ColorJitter and moire pattern augmentation, called SPSC, and achieved the best ACER result of 1.32\% in protocol 2.1. Compared with the baseline, the ACER was reduced by 36.73\%.

 % Using the complete training set and live samples from the development set and employing Resnet50 \cite{he2016deep} as the backbone,

\begin{table}[t]
\footnotesize
\centering
% \resizebox{0.47\textwidth}{10mm}{
% \vspace{2mm}
\begin{tabular}{c c c c c c}
\toprule
    Backbone & Moire & Color & APCER $\downarrow$ & BPCER $\downarrow$ & ACER $\downarrow$ \\ 
    \midrule
    Resnet50 & \XSolidBrush & \XSolidBrush & 76.03  &  0.06  &  38.05 \\
    Resnet50 & \CheckmarkBold & \XSolidBrush & 11.07  &  1.28  &  6.18 \\
    Resnet50 & \XSolidBrush & \CheckmarkBold & 4.74 &  2.51 &  3.62 \\
    Resnet50 & \CheckmarkBold & \CheckmarkBold & 2.55  & 0.09   &  \textbf{1.32} \\
    % w/o DIFF &  x &  x &  x \\      
\bottomrule
\end{tabular}
\caption{Ablation studies. Protocol 2.1 comparisons reveal that the combined moire pattern augmentation and ColorJitter(Color), termed SPSC, excel with the lowest ACER of 1.32\%.}
\label{table:spsc}
\end{table}

% \begin{table}[t]
% % \setlength{\tabcolsep}{4pt}
% \footnotesize
% \centering
% \resizebox{0.47\textwidth}{10mm}{
% % \vspace{2mm}
% \begin{tabular}{c c c c c c}
% \toprule
%     backbone & moire & ColorJitter & APCER $\downarrow$ & BPCER $\downarrow$ & ACER $\downarrow$ \\ \midrule
%     Resnet50 & \XSolidBrush & \XSolidBrush & 76.03  &  0.06  &  38.05 \\
%     Resnet50 & \CheckmarkBold & \XSolidBrush & 11.07  &  1.28  &  6.18 \\
%     Resnet50 & \XSolidBrush & \CheckmarkBold & 4.74 &  2.51 &  3.62 \\
%     Resnet50 & \CheckmarkBold & \CheckmarkBold & 0.09  &  2.55  &  \textbf{1.32} \\
%     % w/o DIFF &  x &  x &  x \\      
% \bottomrule
% \end{tabular}}
% \caption{Ablation studies. Protocol 2.1 comparisons reveal that the combined moire pattern augmentation and ColorJitter, termed SPSC, excel with the lowest ACER of 1.32\%.}
% \label{table:spsc}
% \end{table}

\paragraph{Simulated Digital Spoofing Clues} 
SDSC is developed to simulate spoofing cues from digital attacks. As shown in Table \ref{table:sdsc}, using the complete training set and live samples from the development set and employing Resnet50 \cite{he2016deep} as the backbone, SDSC significantly lowers the baseline ACER from 44.35\% to 1.65\%, a reduction of 42.7\%, underscoring the effectiveness of SDSC. Conversely, GaussNoise, intended to mimic adversarial attack cues, was less effective, resulting in an ACER of 36.77\%. The combined use of SDSC and GaussNoise yields an ACER of 22.57\%.
% SDSC are used to simulate the spoofing clues of digital attack. In protocol 2.2, the full train set and live samples in the dev set are used for training, and Resnet50\cite{he2016deep} is used as the backbone to train the baseline result. As presented in Table \ref{table:sdsc}, compared to the baseline ACER result of 44.35\%, using SDSC achieved the best ACER of 1.65\%, and the ACER was reduced by 42.7\%. ACER has been significantly reduced, indicating that the SDSC is very effective. We also used GaussNoise to simulate the spoofing clues of adversarial attacks whose spoofing clues are gradient noise, but it did not work, with an ACER of 48.42\%.

\begin{table}[t]
\footnotesize
\centering
% \vspace{2mm}
% \resizebox{0.47\textwidth}{10mm}{
% \begin{tabularx}{0.47\textwidth}{X X X X X X}
\begin{tabular}{c c c c c c}
\toprule
    Backbone & SDSC & Noise & APCER $\downarrow$ & BPCER $\downarrow$ & ACER $\downarrow$ \\ 
    \midrule
    Resnet50 & \XSolidBrush & \XSolidBrush & 88.59  &  0.11  &  44.35 \\
    Resnet50 & \CheckmarkBold & \XSolidBrush & 1.73  &  1.58  &  \textbf{1.65} \\
    Resnet50 & \XSolidBrush & \CheckmarkBold & 71.60 &  1.93 &  36.77 \\
    Resnet50 & \CheckmarkBold & \CheckmarkBold & 44.0 & 1.08 &  22.57 \\
    % w/o DIFF &  x &  x &  x \\      
\bottomrule
\end{tabular}
\caption{In Protocol 2.2, ablation studies contrast the effects of SDSC and GaussNoise(Noise). Employing only SDSC achieved an optimal ACER of 1.65\%. In contrast, GaussNoise, designed to simulate adversarial attack noise, proved ineffective, resulting in an ACER of 36.77\%.}
\label{table:sdsc}
\end{table}

\subsection{Comparisons of Different Backbones}
As shown in Table \ref{table:backbone}, we show the performance of our method with five different backbones: Resnet18, Resnet34, Resnet50~\cite{he2016deep}, Swin-Tiny and Swin-Base~\cite{swin}. Due to the smaller size of the training set in the UniAttackData Dataset~\cite{fang2024unified}, our method exhibited some fluctuations in results across different backbones. However, these fluctuations remained within a reasonable range. We posit that our approach functions akin to a plug-in, capable of being easily integrated with any backbone architecture. This adaptability suggests that our method is not tightly coupled with the network structure. Our belief is that the proposed method is agnostic to the network structure, which is a significant advantage for its application in various scenarios. Ultimately, our method achieved the best ACER result of 1.06\% on the Resnet50 backbone, demonstrating its effectiveness and the potential for broader applicability.

\begin{table}[t]
\footnotesize
\centering
% \vspace{2mm}
\begin{tabular}{c c c c c}
\toprule
    Backbone & AUC $\uparrow$ & APCER $\downarrow$ & BPCER $\downarrow$ & ACER $\downarrow$ \\ 
    \midrule
    Resnet18~\cite{he2016deep} &  99.70 & 3.19  & 1.41   &  2.30 \\
    Resnet34~\cite{he2016deep} &  99.83  &  2.33  & 1.03   &  1.68 \\
    Resnet50~\cite{he2016deep} &  99.94  &  1.53 &  0.69 &  1.06 \\
    Swin-Tiny~\cite{swin} &  99.88  &  2.10 &  0.96  &  1.53 \\
    Swin-Base~\cite{swin} &  99.79  &  3.15 &  1.27 &  2.21 \\   
\bottomrule
\end{tabular}
\caption{Comparison of our method with different backbones.}
\label{table:backbone}
\end{table}

\begin{figure}[t]
     \centering
     \includegraphics[width=\linewidth]{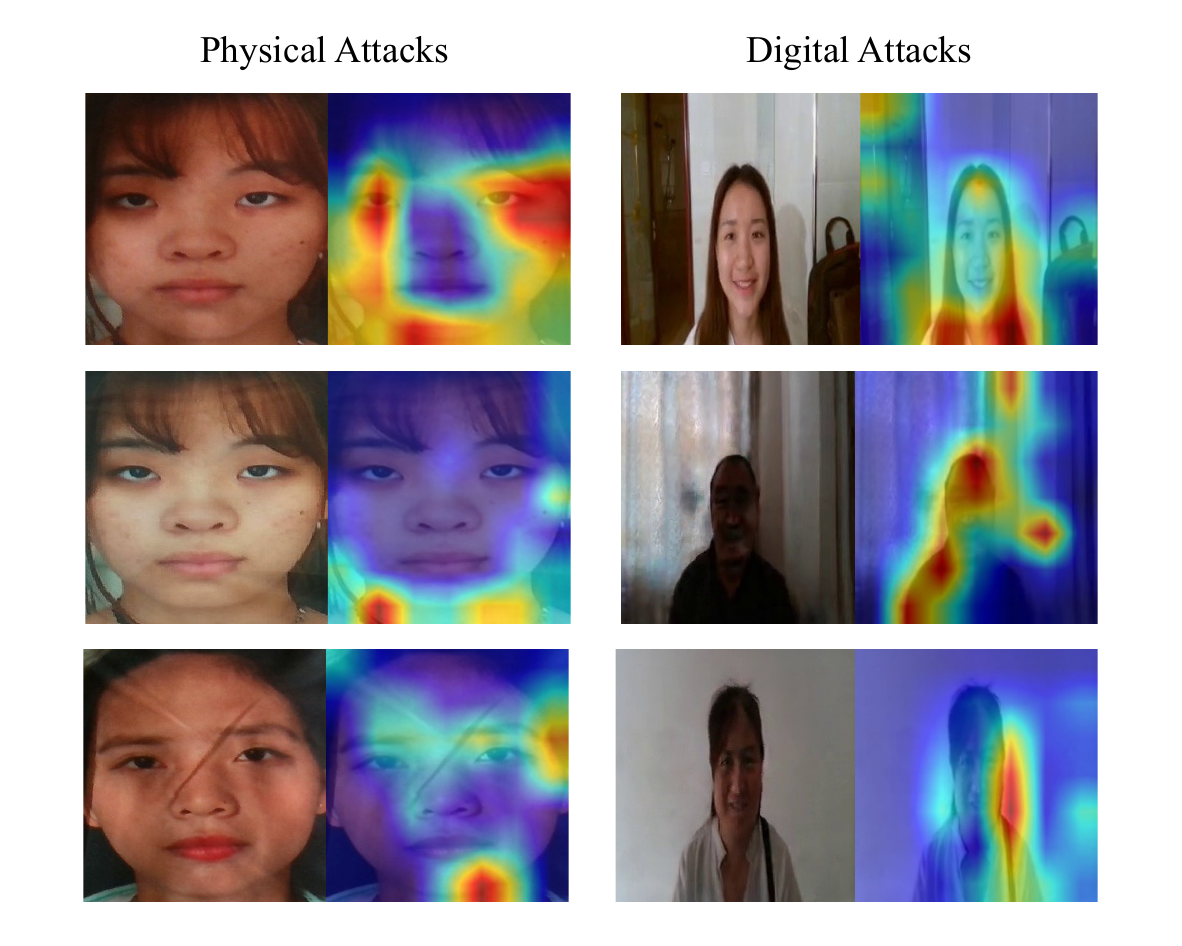}
     \caption{Visualization of attention maps for physical attacks and digital attacks.}
     \label{fig:view}
    % \vspace{-0.5cm}
\end{figure}

\subsection{Visualizations}
% \paragraph{Visual Results} 
We employ GradCam~\cite{jacobgilpytorchcam} to interpret and locate attack-specific features within the challenge. Figure \ref{fig:view} illustrates that in Protocol 2.1, the focus of the model is mainly on areas with color distortion or moire patterns, indicative of physical attacks. In Protocol 2.2, attention shifts to facial edge artifacts or distortion regions associated with digital attacks. These patterns confirm that our model has effectively learned to identify spoofing cues for both physical and digital attacks, achieving superior performance.

% We use GradCam\cite{jacobgilpytorchcam}, a gradient localization-based visual interpretation technique, to analyze different types of attacks in this challenge. As shown in Figure \ref{fig:view}, in protocol 2.1, for physical attack types, attention is focused more on the color distortion area or moire pattern area of the image. This shows that the model we trained learned the spoofing clues of physical attacks achieved advanced performance. In Protocol 2.2, for digital attack types, attention is focused more on the artifacts of face edge contours or distortion areas of the image. This shows that the model we trained learned the spoofing clues of digital attacks achieved advanced performance.

% \vspace{-0.2cm}
\section{Conclusion}
% \vspace{-0.15cm}
In this paper, our method introduces two novel data augmentations: Simulated Physical Spoofing Clues augmentation (SPSC) and Simulated Digital Spoofing Clues augmentation (SDSC). Extensive experimentation demonstrates their substantial enhancement of the model's detection and generalization capabilities for ``unseen" attacks. Finally, our method won first place in ``Unified Physical-Digital Face Attack Detection" of the 5th Face Anti-spoofing Challenge@CVPR2024.

% In this paper, our method proposes two data augmentation methods called simulate physical spoofing clues augmentation(SPSC) and simulate digital spoofing clues augmentation(SDSC). Extensive experiments show that our method greatly improves the model's detection ability and generalization of "unseen" attack types. Finally, our method won the first place in "Unified Physical-Digital Face Attack Detection" of the 5th Face Anti-spoofing Challenge@CVPR2024.

{
    \small
    \bibliographystyle{ieeenat_fullname}
    \bibliography{main}

\begin{thebibliography}{53}
\providecommand{\natexlab}[1]{#1}
\providecommand{\url}[1]{\texttt{#1}}
\expandafter\ifx\csname urlstyle\endcsname\relax
  \providecommand{\doi}[1]{doi: #1}\else
  \providecommand{\doi}{doi: \begingroup \urlstyle{rm}\Url}\fi

\bibitem[Chen et~al.(2020)Chen, Chen, Ni, and Ge]{chen2020simswap}
Renwang Chen, Xuanhong Chen, Bingbing Ni, and Yanhao Ge.
\newblock Simswap: An efficient framework for high fidelity face swapping.
\newblock In \emph{Proceedings of the 28th ACM international conference on multimedia}, pages 2003--2011, 2020.

\bibitem[Chen et~al.(2021)Chen, Yao, Sheng, Ding, Tai, Li, Huang, and Jin]{Chen2021a}
Zhihong Chen, Taiping Yao, Kekai Sheng, Shouhong Ding, Ying Tai, Jilin Li, Feiyue Huang, and Xinyu Jin.
\newblock Generalizable representation learning for mixture domain face anti-spoofing.
\newblock In \emph{AAAI}, 2021.

\bibitem[Duan et~al.(2021)Duan, Chen, Niu, Yang, Qin, and He]{duan2021advdrop}
Ranjie Duan, Yuefeng Chen, Dantong Niu, Yun Yang, A~Kai Qin, and Yuan He.
\newblock Advdrop: Adversarial attack to dnns by dropping information.
\newblock In \emph{Proceedings of the IEEE/CVF International Conference on Computer Vision}, pages 7506--7515, 2021.

\bibitem[Fang et~al.(2023)Fang, Liu, Wan, Escalera, Zhao, Zhang, Li, and Lei]{fang2023surveillance}
Hao Fang, Ajian Liu, Jun Wan, Sergio Escalera, Chenxu Zhao, Xu Zhang, Stan~Z Li, and Zhen Lei.
\newblock Surveillance face anti-spoofing.
\newblock \emph{IEEE Transactions on Information Forensics and Security}, 2023.

\bibitem[Fang et~al.(2024)Fang, Liu, Yuan, Zheng, Zeng, Liu, Deng, Escalera, Liu, Wan, et~al.]{fang2024unified}
Hao Fang, Ajian Liu, Haocheng Yuan, Junze Zheng, Dingheng Zeng, Yanhong Liu, Jiankang Deng, Sergio Escalera, Xiaoming Liu, Jun Wan, et~al.
\newblock Unified physical-digital face attack detection.
\newblock \emph{arXiv preprint arXiv:2401.17699}, 2024.

\bibitem[Feng et~al.(2020)Feng, Hong, Yue, Chen, Wang, Han, Liu, and Ding]{feng2020learning}
Haocheng Feng, Zhibin Hong, Haixiao Yue, Yang Chen, Keyao Wang, Junyu Han, Jingtuo Liu, and Errui Ding.
\newblock Learning generalized spoof cues for face anti-spoofing.
\newblock \emph{arXiv preprint arXiv:2005.03922}, 2020.

\bibitem[Gildenblat and contributors(2021)]{jacobgilpytorchcam}
Jacob Gildenblat and contributors.
\newblock Pytorch library for cam methods.
\newblock \url{https://github.com/jacobgil/pytorch-grad-cam}, 2021.

\bibitem[Guo et~al.(2023)Guo, Zhen, and Yan]{guo2023controllable}
Ying Guo, Cheng Zhen, and Pengfei Yan.
\newblock Controllable guide-space for generalizable face forgery detection.
\newblock In \emph{Proceedings of the IEEE/CVF International Conference on Computer Vision}, pages 20818--20827, 2023.

\bibitem[He et~al.(2016)He, Zhang, Ren, and Sun]{he2016deep}
Kaiming He, Xiangyu Zhang, Shaoqing Ren, and Jian Sun.
\newblock Deep residual learning for image recognition.
\newblock In \emph{Proceedings of the IEEE conference on computer vision and pattern recognition}, pages 770--778, 2016.

\bibitem[Heusch et~al.(2020)Heusch, George, Geissb{\"u}hler, Mostaani, and Marcel]{heusch2020deep}
Guillaume Heusch, Anjith George, David Geissb{\"u}hler, Zohreh Mostaani, and S{\'e}bastien Marcel.
\newblock Deep models and shortwave infrared information to detect face presentation attacks.
\newblock \emph{IEEE Transactions on Biometrics, Behavior, and Identity Science}, 2\penalty0 (4):\penalty0 399--409, 2020.

\bibitem[Hong et~al.(2022)Hong, Zhang, Shen, and Xu]{hong2022depth}
Fa-Ting Hong, Longhao Zhang, Li Shen, and Dan Xu.
\newblock Depth-aware generative adversarial network for talking head video generation.
\newblock In \emph{Proceedings of the IEEE/CVF conference on computer vision and pattern recognition}, pages 3397--3406, 2022.

\bibitem[Huang et~al.(2022)Huang, Sun, Liu, Chu, Xiao, Yuan, Adam, and Yang]{huang2022adaptive}
Hsin-Ping Huang, Deqing Sun, Yaojie Liu, Wen-Sheng Chu, Taihong Xiao, Jinwei Yuan, Hartwig Adam, and Ming-Hsuan Yang.
\newblock Adaptive transformers for robust few-shot cross-domain face anti-spoofing.
\newblock \emph{arXiv preprint arXiv:2203.12175}, 2022.

\bibitem[Jia et~al.(2020{\natexlab{a}})Jia, Guo, and Xu]{jia2020survey}
Shan Jia, Guodong Guo, and Zhengquan Xu.
\newblock A survey on 3d mask presentation attack detection and countermeasures.
\newblock \emph{Pattern recognition}, 98:\penalty0 107032, 2020{\natexlab{a}}.

\bibitem[Jia et~al.(2020{\natexlab{b}})Jia, Zhang, Shan, and Chen]{jia2020single}
Yunpei Jia, Jie Zhang, Shiguang Shan, and Xilin Chen.
\newblock Single-side domain generalization for face anti-spoofing.
\newblock In \emph{Proceedings of the IEEE/CVF Conference on Computer Vision and Pattern Recognition}, pages 8484--8493, 2020{\natexlab{b}}.

\bibitem[Kim et~al.(2021)Kim, Tariq, and Woo]{kim2021fretal}
Minha Kim, Shahroz Tariq, and Simon~S Woo.
\newblock Fretal: Generalizing deepfake detection using knowledge distillation and representation learning.
\newblock In \emph{Proceedings of the IEEE/CVF conference on computer vision and pattern recognition}, pages 1001--1012, 2021.

\bibitem[Li et~al.(2021)Li, Yuan, Chen, Wang, and Vasconcelos]{li2021dynamic}
Yunsheng Li, Lu Yuan, Yinpeng Chen, Pei Wang, and Nuno Vasconcelos.
\newblock Dynamic transfer for multi-source domain adaptation.
\newblock In \emph{Proceedings of the IEEE/CVF Conference on Computer Vision and Pattern Recognition}, pages 10998--11007, 2021.

\bibitem[Liu and Liang(2022)]{ijcai2022p165}
Ajian Liu and Yanyan Liang.
\newblock Ma-vit: Modality-agnostic vision transformers for face anti-spoofing.
\newblock In \emph{Proceedings of the Thirty-First International Joint Conference on Artificial Intelligence, {IJCAI-22}}, pages 1180--1186, 2022.

\bibitem[Liu et~al.(2019)Liu, Wan, Escalera, Jair~Escalante, Tan, Yuan, Wang, Lin, Guo, Guyon, et~al.]{liu2019multi}
Ajian Liu, Jun Wan, Sergio Escalera, Hugo Jair~Escalante, Zichang Tan, Qi Yuan, Kai Wang, Chi Lin, Guodong Guo, Isabelle Guyon, et~al.
\newblock Multi-modal face anti-spoofing attack detection challenge at cvpr2019.
\newblock In \emph{Proceedings of the IEEE/CVF conference on computer vision and pattern recognition workshops}, pages 0--0, 2019.

\bibitem[Liu et~al.(2021{\natexlab{a}})Liu, Li, Wan, Liang, Escalera, Escalante, Madadi, Jin, Wu, Yu, et~al.]{liu2021cross}
Ajian Liu, Xuan Li, Jun Wan, Yanyan Liang, Sergio Escalera, Hugo~Jair Escalante, Meysam Madadi, Yi Jin, Zhuoyuan Wu, Xiaogang Yu, et~al.
\newblock Cross-ethnicity face anti-spoofing recognition challenge: A review.
\newblock \emph{IET Biometrics}, 10\penalty0 (1):\penalty0 24--43, 2021{\natexlab{a}}.

\bibitem[Liu et~al.(2021{\natexlab{b}})Liu, Tan, Wan, Escalera, Guo, and Li]{liu2021casia}
Ajian Liu, Zichang Tan, Jun Wan, Sergio Escalera, Guodong Guo, and Stan~Z Li.
\newblock Casia-surf cefa: A benchmark for multi-modal cross-ethnicity face anti-spoofing.
\newblock In \emph{Proceedings of the IEEE/CVF Winter Conference on Applications of Computer Vision}, pages 1179--1187, 2021{\natexlab{b}}.

\bibitem[Liu et~al.(2021{\natexlab{c}})Liu, Tan, Wan, Liang, Lei, Guo, and Li]{liu2021face}
Ajian Liu, Zichang Tan, Jun Wan, Yanyan Liang, Zhen Lei, Guodong Guo, and Stan~Z Li.
\newblock Face anti-spoofing via adversarial cross-modality translation.
\newblock \emph{IEEE Transactions on Information Forensics and Security}, 16:\penalty0 2759--2772, 2021{\natexlab{c}}.

\bibitem[Liu et~al.(2021{\natexlab{d}})Liu, Zhao, Yu, Su, Liu, Kong, Wan, Escalera, Escalante, Lei, et~al.]{liu20213d}
Ajian Liu, Chenxu Zhao, Zitong Yu, Anyang Su, Xing Liu, Zijian Kong, Jun Wan, Sergio Escalera, Hugo~Jair Escalante, Zhen Lei, et~al.
\newblock 3d high-fidelity mask face presentation attack detection challenge.
\newblock In \emph{Proceedings of the IEEE/CVF International Conference on Computer Vision Workshops}, pages 814--823, 2021{\natexlab{d}}.

\bibitem[Liu et~al.(2022{\natexlab{a}})Liu, Wan, Jiang, Wang, and Liang]{liu2022disentangling}
Ajian Liu, Jun Wan, Ning Jiang, Hongbin Wang, and Yanyan Liang.
\newblock Disentangling facial pose and appearance information for face anti-spoofing.
\newblock In \emph{2022 26th International Conference on Pattern Recognition (ICPR)}, pages 4537--4543. IEEE, 2022{\natexlab{a}}.

\bibitem[Liu et~al.(2022{\natexlab{b}})Liu, Zhao, Yu, Wan, Su, Liu, Tan, Escalera, Xing, Liang, et~al.]{liu2022contrastive}
Ajian Liu, Chenxu Zhao, Zitong Yu, Jun Wan, Anyang Su, Xing Liu, Zichang Tan, Sergio Escalera, Junliang Xing, Yanyan Liang, et~al.
\newblock Contrastive context-aware learning for 3d high-fidelity mask face presentation attack detection.
\newblock \emph{IEEE Transactions on Information Forensics and Security}, 17:\penalty0 2497--2507, 2022{\natexlab{b}}.

\bibitem[Liu et~al.(2023{\natexlab{a}})Liu, Tan, Liang, and Wan]{Liu_2023_CVPR}
Ajian Liu, Zichang Tan, Yanyan Liang, and Jun Wan.
\newblock Attack-agnostic deep face anti-spoofing.
\newblock In \emph{Proceedings of the IEEE/CVF Conference on Computer Vision and Pattern Recognition (CVPR) Workshops}, pages 6335--6344, 2023{\natexlab{a}}.

\bibitem[Liu et~al.(2023{\natexlab{b}})Liu, Tan, Yu, Zhao, Wan, Lei, Zhang, Li, and Guo]{liu2023fm}
Ajian Liu, Zichang Tan, Zitong Yu, Chenxu Zhao, Jun Wan, Yanyan Liang~Zhen Lei, Du Zhang, Stan~Z Li, and Guodong Guo.
\newblock Fm-vit: Flexible modal vision transformers for face anti-spoofing.
\newblock \emph{IEEE Transactions on Information Forensics and Security}, 2023{\natexlab{b}}.

\bibitem[Liu et~al.(2024)Liu, Xue, Gan, Wan, Liang, Deng, Escalera, and Lei]{liu2024cfplfas}
Ajian Liu, Shuai Xue, Jianwen Gan, Jun Wan, Yanyan Liang, Jiankang Deng, Sergio Escalera, and Zhen Lei.
\newblock Cfpl-fas: Class free prompt learning for generalizable face anti-spoofing.
\newblock In \emph{Proceedings of the IEEE/CVF Conference on Computer Vision and Pattern Recognition}, 2024.

\bibitem[Liu et~al.(2021{\natexlab{e}})Liu, Zhang, Yao, Bi, Ding, Li, Huang, and Ma]{liu2021adaptive}
Shubao Liu, Ke-Yue Zhang, Taiping Yao, Mingwei Bi, Shouhong Ding, Jilin Li, Feiyue Huang, and Lizhuang Ma.
\newblock Adaptive normalized representation learning for generalizable face anti-spoofing.
\newblock In \emph{Proceedings of the 29th ACM International Conference on Multimedia}, pages 1469--1477, 2021{\natexlab{e}}.

\bibitem[Liu et~al.(2018)Liu, Jourabloo, and Liu]{Liu2018Learning}
Yaojie Liu, Amin Jourabloo, and Xiaoming Liu.
\newblock Learning deep models for face anti-spoofing: Binary or auxiliary supervision.
\newblock In \emph{CVPR}, 2018.

\bibitem[Liu et~al.(2021{\natexlab{f}})Liu, Lin, Cao, Hu, Wei, Zhang, Lin, and Guo]{swin}
Ze Liu, Yutong Lin, Yue Cao, Han Hu, Yixuan Wei, Zheng Zhang, Stephen Lin, and Baining Guo.
\newblock Swin transformer: Hierarchical vision transformer using shifted windows.
\newblock In \emph{Proceedings of the IEEE/CVF international conference on computer vision}, pages 10012--10022, 2021{\natexlab{f}}.

\bibitem[Loshchilov and Hutter(2017)]{adamw}
Ilya Loshchilov and Frank Hutter.
\newblock Decoupled weight decay regularization.
\newblock \emph{arXiv preprint arXiv:1711.05101}, 2017.

\bibitem[Luo et~al.(2022)Luo, Lin, Xie, Wu, Xie, and Shen]{luo2022frequency}
Cheng Luo, Qinliang Lin, Weicheng Xie, Bizhu Wu, Jinheng Xie, and Linlin Shen.
\newblock Frequency-driven imperceptible adversarial attack on semantic similarity.
\newblock In \emph{Proceedings of the IEEE/CVF conference on computer vision and pattern recognition}, pages 15315--15324, 2022.

\bibitem[Nadimpalli and Rattani(2022)]{nadimpalli2022improving}
Aakash~Varma Nadimpalli and Ajita Rattani.
\newblock On improving cross-dataset generalization of deepfake detectors.
\newblock In \emph{Proceedings of the IEEE/CVF conference on computer vision and pattern recognition}, pages 91--99, 2022.

\bibitem[Rony et~al.(2021)Rony, Granger, Pedersoli, and Ben~Ayed]{rony2021augmented}
J{\'e}r{\^o}me Rony, Eric Granger, Marco Pedersoli, and Ismail Ben~Ayed.
\newblock Augmented lagrangian adversarial attacks.
\newblock In \emph{Proceedings of the IEEE/CVF International Conference on Computer Vision}, pages 7738--7747, 2021.

\bibitem[Rosberg et~al.(2023)Rosberg, Aksoy, Alonso-Fernandez, and Englund]{rosberg2023facedancer}
Felix Rosberg, Eren~Erdal Aksoy, Fernando Alonso-Fernandez, and Cristofer Englund.
\newblock Facedancer: Pose-and occlusion-aware high fidelity face swapping.
\newblock In \emph{Proceedings of the IEEE/CVF winter conference on applications of computer vision}, pages 3454--3463, 2023.

\bibitem[Shao et~al.(2019)Shao, Lan, Li, and Yuen]{shao2019multi}
Rui Shao, Xiangyuan Lan, Jiawei Li, and Pong~C Yuen.
\newblock Multi-adversarial discriminative deep domain generalization for face presentation attack detection.
\newblock In \emph{Proceedings of the IEEE/CVF conference on computer vision and pattern recognition}, pages 10023--10031, 2019.

\bibitem[Shiohara and Yamasaki(2022)]{shiohara2022detecting}
Kaede Shiohara and Toshihiko Yamasaki.
\newblock Detecting deepfakes with self-blended images.
\newblock In \emph{Proceedings of the IEEE/CVF Conference on Computer Vision and Pattern Recognition}, pages 18720--18729, 2022.

\bibitem[Srivatsan et~al.(2023)Srivatsan, Naseer, and Nandakumar]{srivatsan1}
Koushik Srivatsan, Muzammal Naseer, and Karthik Nandakumar.
\newblock Flip: Cross-domain face anti-spoofing with language guidance.
\newblock In \emph{ICCV}, 2023.

\bibitem[Sun et~al.(2023)Sun, Liu, Liu, Li, and Chu]{sun2023rethinking}
Yiyou Sun, Yaojie Liu, Xiaoming Liu, Yixuan Li, and Wen-Sheng Chu.
\newblock Rethinking domain generalization for face anti-spoofing: Separability and alignment.
\newblock In \emph{Proceedings of the IEEE/CVF Conference on Computer Vision and Pattern Recognition}, pages 24563--24574, 2023.

\bibitem[Wang et~al.(2024)Wang, Zhang, Yue, Liu, Zhang, Feng, Han, Ding, and Wang]{wang2024multi}
Keyao Wang, Guosheng Zhang, Haixiao Yue, Ajian Liu, Gang Zhang, Haocheng Feng, Junyu Han, Errui Ding, and Jingdong Wang.
\newblock Multi-domain incremental learning for face presentation attack detection.
\newblock In \emph{Proceedings of the AAAI Conference on Artificial Intelligence}, pages 5499--5507, 2024.

\bibitem[Wang et~al.(2021{\natexlab{a}})Wang, Zhang, and Li]{wang2021safa}
Qiulin Wang, Lu Zhang, and Bo Li.
\newblock Safa: Structure aware face animation.
\newblock In \emph{2021 International Conference on 3D Vision (3DV)}, pages 679--688. IEEE, 2021{\natexlab{a}}.

\bibitem[Wang et~al.(2021{\natexlab{b}})Wang, Mallya, and Liu]{wang2021one}
Ting-Chun Wang, Arun Mallya, and Ming-Yu Liu.
\newblock One-shot free-view neural talking-head synthesis for video conferencing.
\newblock In \emph{Proceedings of the IEEE/CVF conference on computer vision and pattern recognition}, pages 10039--10049, 2021{\natexlab{b}}.

\bibitem[Wang et~al.(2021{\natexlab{c}})Wang, Wu, Jiang, Hao, Tan, and Zhang]{wang2021demiguise}
Yajie Wang, Shangbo Wu, Wenyi Jiang, Shengang Hao, Yu-an Tan, and Quanxin Zhang.
\newblock Demiguise attack: Crafting invisible semantic adversarial perturbations with perceptual similarity.
\newblock \emph{arXiv preprint arXiv:2107.01396}, 2021{\natexlab{c}}.

\bibitem[Yan et~al.(2022)Yan, Cheung, and Yeung]{yan2022ila}
Chiu~Wai Yan, Tsz-Him Cheung, and Dit-Yan Yeung.
\newblock Ila-da: Improving transferability of intermediate level attack with data augmentation.
\newblock In \emph{The Eleventh International Conference on Learning Representations}, 2022.

\bibitem[Yu et~al.(2021)Yu, Zhu, Chen, Zhao, Chen, Tang, Zhu, and Qiao]{yu2021multiple}
Shijie Yu, Feng Zhu, Dapeng Chen, Rui Zhao, Haobin Chen, Shixiang Tang, Jinguo Zhu, and Yu Qiao.
\newblock Multiple domain experts collaborative learning: Multi-source domain generalization for person re-identification.
\newblock \emph{arXiv preprint arXiv:2105.12355}, 2021.

\bibitem[Yu et~al.(2020)Yu, Zhao, Wang, Qin, Su, Li, Zhou, and Zhao]{yu2020searching}
Zitong Yu, Chenxu Zhao, Zezheng Wang, Yunxiao Qin, Zhuo Su, Xiaobai Li, Feng Zhou, and Guoying Zhao.
\newblock Searching central difference convolutional networks for face anti-spoofing.
\newblock In \emph{Proceedings of the IEEE/CVF conference on computer vision and pattern recognition}, pages 5295--5305, 2020.

\bibitem[Yu et~al.(2023{\natexlab{a}})Yu, Cai, Cui, Liu, and Chen]{yu2023visual}
Zitong Yu, Rizhao Cai, Yawen Cui, Ajian Liu, and Changsheng Chen.
\newblock Visual prompt flexible-modal face anti-spoofing, 2023{\natexlab{a}}.

\bibitem[Yu et~al.(2023{\natexlab{b}})Yu, Liu, Zhao, Cheng, Cheng, and Zhao]{yu2023flexiblemodal}
Zitong Yu, Ajian Liu, Chenxu Zhao, Kevin H.~M. Cheng, Xu Cheng, and Guoying Zhao.
\newblock Flexible-modal face anti-spoofing: A benchmark, 2023{\natexlab{b}}.

\bibitem[Zhang et~al.(2019)Zhang, Wang, Liu, Zhao, Wan, Escalera, Shi, Wang, and Li]{zhang2019dataset}
Shifeng Zhang, Xiaobo Wang, Ajian Liu, Chenxu Zhao, Jun Wan, Sergio Escalera, Hailin Shi, Zezheng Wang, and Stan~Z Li.
\newblock A dataset and benchmark for large-scale multi-modal face anti-spoofing.
\newblock In \emph{Proceedings of the IEEE/CVF Conference on Computer Vision and Pattern Recognition}, pages 919--928, 2019.

\bibitem[Zhang et~al.(2020)Zhang, Liu, Wan, Liang, Guo, Escalera, Escalante, and Li]{zhang2020casia}
Shifeng Zhang, Ajian Liu, Jun Wan, Yanyan Liang, Guodong Guo, Sergio Escalera, Hugo~Jair Escalante, and Stan~Z Li.
\newblock Casia-surf: A large-scale multi-modal benchmark for face anti-spoofing.
\newblock \emph{IEEE Transactions on Biometrics, Behavior, and Identity Science}, 2\penalty0 (2):\penalty0 182--193, 2020.

\bibitem[Zhou et~al.(2023)Zhou, Zhang, Yao, Lu, Yi, Ding, and Ma]{zhou2023instance}
Qianyu Zhou, Ke-Yue Zhang, Taiping Yao, Xuequan Lu, Ran Yi, Shouhong Ding, and Lizhuang Ma.
\newblock Instance-aware domain generalization for face anti-spoofing.
\newblock In \emph{Proceedings of the IEEE/CVF Conference on Computer Vision and Pattern Recognition}, pages 20453--20463, 2023.

\bibitem[Zhuang et~al.(2022)Zhuang, Chu, Tan, Liu, Yuan, Miao, Luo, and Yu]{zhuang2022uia}
Wanyi Zhuang, Qi Chu, Zhentao Tan, Qiankun Liu, Haojie Yuan, Changtao Miao, Zixiang Luo, and Nenghai Yu.
\newblock Uia-vit: Unsupervised inconsistency-aware method based on vision transformer for face forgery detection.
\newblock In \emph{European Conference on Computer Vision}, pages 391--407. Springer, 2022.

\bibitem[Zou et~al.(2022)Zou, Duan, Li, Zhang, Pan, and Pan]{zou2022making}
Junhua Zou, Yexin Duan, Boyu Li, Wu Zhang, Yu Pan, and Zhisong Pan.
\newblock Making adversarial examples more transferable and indistinguishable.
\newblock In \emph{Proceedings of the AAAI Conference on Artificial Intelligence}, pages 3662--3670, 2022.

\end{thebibliography}
}

\end{document}